\documentclass[a4paper, 12pt]{apa}
\usepackage[utf8]{inputenc}
\usepackage{authblk}
\usepackage{setspace}
\usepackage{graphicx}
\usepackage{float}
\usepackage{listings}
\usepackage{natbib}
\usepackage{color}
\usepackage[normalem]{ulem}
\definecolor{ultramarine}{RGB}{0,32,96}
\usepackage{booktabs}
\usepackage{listings}
\usepackage{sverb}

\newcommand{\delete}[1]{}

\newcommand{\comment}[2]{}

\title{Modeling VI and VDRL feedback functions: Searching normative rules through computational simulation}
\date{}

\author[1,2]{Paulo Sergio Panse Silveira}
\author[2]{José de Oliveira Siqueira}
\author[3,4]{João Lucas Bernardy}
\author[3]{Jessica Santiago}
\author[3]{Thiago Cersosimo Meneses}
\author[3]{Bianca Sanches Portela}
\author[3,4]{Marcelo Frota Benvenuti*}
\affil[1]{Department of Pathology, Medical School, University of Sao Paulo, São Paulo, SP, Brazil}
\affil[2]{Department of Legal Medicine, Bioethics, Occupational Medicine and Physical Medicine and Rehabilitation, Medical School, University of Sao Paulo, São Paulo, SP, Brazil}
\affil[3]{Department of Experimental Psychology, Institute of Psychology, University of Sao Paulo, São Paulo, SP, Brazil}
\affil[4]{National Institute of Science and Technology: Behavior, Cognition and Teaching (INCT-ECCE)}

\begin{document}

\maketitle

\textbf{* corresponding author}

mbenvenuti@usp.br\newline
Instituto de Psicologia da USP\newline
Av. Prof. Mello Moraes, 1721\newline
05508-030, Sao Paulo, SP, Brazil

\clearpage
\tableofcontents


\clearpage

\section{ORCID}

Paulo S. P. Silveira: 0000-0003-4110-1038

José O. Siqueira: 0000-0002-3357-8939

João L. Bernardy: 0000-0002-3805-7366

Jessica Santiago: 0000-0002-7788-5455

Thiago C. Meneses: 0000-0003-3473-5841

Bianca S. Portela: 0000-0002-1351-652X

Marcelo F. Benvenuti: 0000-0002-9397-3033

\section{Running head}\label{running-head}

Simple schedules

\section{Keywords}

\begin{itemize}
    \item simulation
    \item simple schedules
\end{itemize}

\section{Data and R scripts}

\url{https://sourceforge.net/projects/simpleschedules}

\clearpage
\singlespacing

\subsection{Author's note}

The authors certify that they have NO affiliations with or involvement in any organization or entity with any financial interest (such as honoraria; educational grants; participation in speakers’ bureaus; membership, employment, consultancies, stock ownership, or other equity interest; and expert testimony or patent-licensing arrangements), or non-financial interest (such as personal or professional relationships, affiliations, knowledge or beliefs) in the subject matter or materials discussed in this manuscript. 

The authors read and approved the final version of the manuscript.

This investigation is purely theoretical, thus it was not submitted to any ethics committee. There are no competing interests to declare.

This manuscript is not under consideration for publication and its individual parts were not published under peer-review journals elsewhere.



\doublespacing

\clearpage
\section{Abstract}\label{abstract}

In this paper, we present a R script named Beak, built to simulate rates of behavior interacting with schedules of reinforcement. Using Beak, we’ve simulated data that allows an assessment of different reinforcement feedback functions (RFF). This was made with unparalleled precision, since simulations provide huge samples of data and, more importantly, simulated behavior isn’t changed by the reinforcement it produces. Therefore, we can vary it systematically. We’ve compared different RFF for RI schedules, using as criteria: meaning, precision, parsimony and generality. Our results indicate that the best feedback function for the RI schedule was published by Baum (1981). We also propose that the model used by Killeen (1975) is a viable feedback function for the RDRL schedule. We argue that Beak paves the way for greater understanding of schedules of reinforcement, addressing still open questions about quantitative features of schedules. Also, they could guide future experiments that use schedules as theoretical and methodological tools.

\clearpage

The general definition of operant behavior implies that behavior controls environmental changes. \citet{Ferster1957} emphasized how these changes shaped different patterns of behavior. For them, behavior was the dependent variable and reinforcement was the independent variable. On the other hand, rate of reinforcement may be treated as the dependent variable and rates of behavior as the independent variable. The mathematical description of such a relation is called Reinforcement feedback function, RFF~\citep{Baum1973,Rachlin1978}. 

Recent technologies pave the way for a more precise quantitative description of reinforcement processes and procedures. A quantitative analysis such as feedback functions and their main features would directly address some old yet still pending questions about schedules of reinforcement \citep[e.g.,][]{Baum1973, Baum1993, Catania1968, Rachlin1978, Killeen1975} and guide future research that uses schedules as a methodological tool. 

In this work, we aim to resume the long-dormant discussion about RFF of simple schedules through a computational routine called Beak. This routine simulates rates of behavior interacting with schedules of reinforcement. Our major contribution is that it allows us to test insurmountable possibilities of rates of responses without having to rely on extensive experimentation with actual subjects. Despite the name Beak, it is important to emphasize that we do not aim to simulate response patterns of any specific animal (e.g., rats, bees, pigeons, humans). Our goal is to build rules about possible outputs of a schedule over a large range of random response rates. These rules could guide future experiments that use schedules as theoretical and methodological tools   

\subsection{Schedules as algorithms}

Schedules of reinforcement are core concepts for the experimental analysis of behavior. The algorithms and rules that define schedules, however, are usually taken for granted, except for initial works \citep[e.g.,][]{Catania1968,Ferster1957,Fleshler1962,Millenson1963}. The absence of schedule appraisal in the current literature is a potential problem, since it could hinder replication.  

A schedule of reinforcement is a set of rules that describe how behavior can produce reinforcers \citep{Ferster1957}. Although the literature on the topic presents a myriad of schedule designations, all of them derive from the criteria used to define the so-called simple schedules. Fundamentally, reinforcers can be a function of a number of responses, of the passage of time, or some combination of both. 

All these schedule requirements can be either fixed (F) or variable (V). On fixed schedules, the criterion to be met (schedule size) is constant between reinforcers. On variable schedules, this criterion is an average of a set of values. Back in the late fifties, implementing a variable schedule could be a challenge. \citet{Ferster1957} did so, selecting a series of values with an intended mean and “scrambling” them. However simple, this solution raises some important questions. How many values should one use? How should the relative frequency of such intervals be distributed? Does scrambling mean randomness? 

Intuitively, one should build a schedule with as many values as possible in order to diminish predictability. Yet, in the past, researchers implemented schedules using a punched tape, in which the distances between holes corresponded to multiples of values that originated the variable schedule. Therefore, this method imposed a practical limitation, because too many values meant very long tapes, which could lead to more technical difficulties \citep{Catania1968}. The electromechanical apparatus also constrained choices regarding the distribution of frequency of interval values. Since it limited the number of values, distributions were always discrete. Instead of variable schedules, modern computers can easily apply random (R) procedures with intervals distributed according to continuous density probability functions as a feasible alternative to fixed and variable schedules.

The absence of discussions addressing the schedule’s algorithms used along many experiments suggests an apparent, but false, consensus. There are several critical aspects to defining and implementing schedules of reinforcement, which were already recognized by \citeauthor{Ferster1957} in their seminal work. According to these authors, every schedule of reinforcement could be “represented by a certain arrangement of timers, counters and relay circuits” \citep[p.~10]{Ferster1957}. Still, most textbooks and technical papers omit relevant details about schedule algorithms and emphasize the behavioral patterns associated with each simple schedule \citep[e.g.,][]{Catania1968, Mazur2016, Pierce2017}.

This discussion, however, is not confined to solely theoretical matters. Schedules of reinforcement are held as crucial methodological tools for behavioral scientists to analyze many experimental results. The correct interpretation of these results relies on clarity of schedule definitions when applied to problems, such as discrimination learning by the use of multiple schedules \citep{Ferster1957, Weiss1974}, observing behavior and conditioned reinforcement \citep{Wyckoff1969}, choice \citep{Herrnstein1961, Herrnstein1970} by the use of concurrent schedules, self-control \citep{Hachlin1972} by the use of concurrent chained schedules, behavior pharmacology \citep{Dews1962, Reilly2003}, decision making and bias \citep{Fantino1998, Goodie1995}.

\subsection{Schedules as feedback functions}

\begin{figure}
\begin{center}
\includegraphics[width=3.5in]{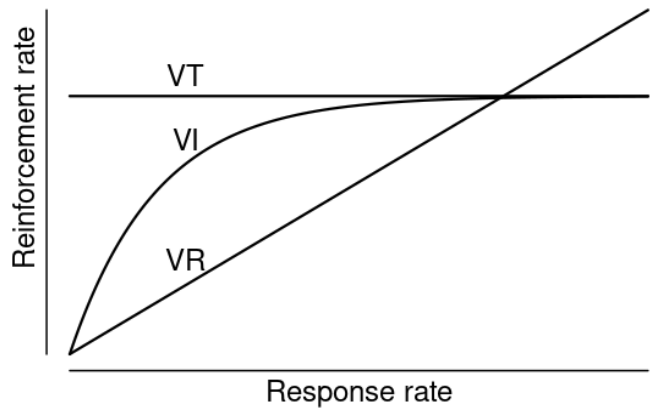}
\end{center}
\caption{Schematic feedback function for three fundamental variable schedules: VT, VI, and VR. VT does not depend on animal behavior for reinforcers are provided at average time intervals. VR completely depends on animal's behavior since reinforcers are provided after a given number of responses. VI is a middle ground, in which the reinforcer becomes available at intervals, but only received after an animal response.}\label{fig:feedback}
\end{figure}

The search for feedback functions for basic schedules is an important attempt to find normative rules about how simple schedules constrain reinforcement. This quantitative signature of schedules precedes the empirical pattern associated with each schedule and the ensuing controversy on differences among species, related repertoires and stability criterion \citep[e.g.,][]{Galizio1988, Stoddard1988}. RFF allows us to discover optimal relations between behavior and reinforcement for each schedule, and so pose a way to propose normative rules for what to expect from actual (optimal) behavior. That is why RFF are a research topic in their own right. Still, the feedback function of many schedules remains an open subject. The general shape of some RFF is well known. Figure \ref{fig:feedback} depicts schematic RFF, based on \citet{Rachlin1989} analysis.

Each RFF clarifies how rates of reinforcement are constrained by basic schedules in a molar level of analysis. Time schedules (e.g., variable time, VT) do not depend on behavior. Therefore, the RFF is a horizontal line with an intercept equal to the rate of reinforcement deduced from the schedule’s size. In ratio schedules (e.g., variable rate, VR), rates of behavior and reinforcement have a linear relation, with an intercept equal to zero and a slope that is the reciprocal of the ratio size. In interval schedules (e.g. variable interval, VI), reinforcement rate is further constrained by a temporal criterion, altering the prior linear function. In such cases, rates of response control increasing rates of reinforcement only until an asymptotic level. 

As \citet{Baum1992} pointed out, a viable RFF should fit the experimental data. But one cannot directly manipulate rates of behavior in the animal laboratory without changing critical aspects of the environment. That’s an important limitation, since RFF models the environment as a function of behavior and the rate of response is the independent variable, but one that we cannot manipulate systematically. Therefore, even with large samples of behavior, experiments rarely cover a sufficiently wide range of response rates \citep{Baum1992}, while simulations allows the experimenter to explore and predict what optimal performances would look like for a wide range of environmental conditions. In fact, experiments with humans and animals do not seem to be the best choice to define a RFF, which was the historical attempt; their utility is the discovery of what strategies among many an organism can adopt and under which circumstances, given the normative rules predicted by simulations~(literally providing a comprehensive map for each schedule), thus opening a whole new string of research.

A consequence of this perception is that the ideal conditions for investigation of RFF are better achievable through computer simulation, because we can prevent simulated behavior from changing as a function of rates of outcomes. Rather we can vary it systematically. For that reason, we have developed Beak, which allows us to analyze with unprecedented precision the quantitative features of feedback functions and build normative rules for different contingencies. 

Although Beak implemented other basic random schedules such as RT and RR, in the present paper, we will discuss the curve fit presented by \citet{Baum1981, Rachlin1978, Prelec1982} for the RI schedule (the pure time or ratio schedules are extremes with monotonous behavior that do not need further discussion for the present focus). We also show that a curve from \citet{Killeen1975}, which was originally proposed in a different theoretical context, is a viable RFF for RI schedules. More interestingly, this function is also a suitable RFF for the random differential reinforcement of low rates (RDRL) that we could extensively map with Beak.

\subsection{Method}

Here we describe how we have implemented simple schedules and responses on Beak. For the sake of parsimony, we will describe the random interval (RI) and random differential reinforcement of low rates (RDRL). The other two basic schedules are simpler and do not pose any fitting challenge: RT is a horizontal line at the schedule size and RR is a simple straight line with slope reciprocal of the ratio size.  Our implementations of simple schedules are mainly based on initial work by \citet{Millenson1963} and \citet{Ambler1973}. We consider their implementation ideal, because they are continuous versions of the discrete (and more widely used) algorithms \citep{Fleshler1962}. Our implementation of responses is like the one by \citet{Green1983}. Distinctly, here, $p$ stands simply for response probability, while $1-p$ stands for a probability of no response at all. Also, trials can happen every fraction of a second, depending on the response rates we want to investigate.

\subsubsection{Random Interval (RI)}

Back in 1963, Millenson proposed the random interval (RI) schedule as a random version of VI schedules \citep{Millenson1963}. Millenson’s RI is a function of the parameters $T$ and $p$, where $T$ stands for the duration of a cycle in any unit of time, at the end of which there is a probability $p$ of reinforcement assignment. The inter-assignment time (IAT) is the number of cycles with duration $T$ until reinforcement assignment. 

For every specific RI size, there are infinite combinations of $T$ and $p$. However, not every combination is eligible for our purposes: behavioral researchers should find values of $T$ and $p$ that will produce an IAT with geometric distribution with mean equal to:

\begin{equation}\label{eq:muRI}
\mu_{RI} = \frac{T}{p}
\end{equation}

In order to achieve a geometric distribution, we must meet two requirements. First, the distribution’s average (Equation \ref{eq:muRI}) must be equal to the standard deviation of the distributions ($\sigma$), given by:

\begin{equation}\label{eq:sigmaRI}
\sigma_{RI} = {\frac{T}{p} \sqrt{1-p}}
\end{equation}

Second, the geometric distribution of IAT will approach an exponential distribution as $T$ approaches zero. The exponential distribution is desirable because it has the inherent property of lack of memory \citep{Feller1968}, \textit{videlicet}, its past behavior bears no information about the future behavior of IAT distribution. This property is key for a more refined implementation of variable schedules because it ensures minimum predictability, as \citet{Fleshler1962} intended. Also, the exponential distribution conveniently portrays the continuity of time. This can be especially useful when using computational simulations, since we have means to investigate exhaustively long sessions with this method. 

On the other hand, \citet{Millenson1963} pointed out that $T$ should also be greater than the average time of reinforcer consumption. For studies with approximately zero consumption time, we argue that $T \le 1$ second is a convenient heuristic for $T$ to meet both requirements simultaneously.

Given that the implemented schedule is a function of $T$ and $p$, it is unlikely that the average and standard deviation will be identical to the planned value. Therefore, we suggest a 1\% margin of tolerance. If $x$ is the planned schedule average and standard deviation, this margin of tolerance for the mean can be described as:

\begin{equation}\label{eq:xplanned}
\frac{\left| x - \frac{T}{p} \right|}{x}  \le {0.01}
\end{equation}

Applying the same margin of tolerance to the standard deviation: 

\begin{equation}\label{eq:xplannedmargin}
1 \ge \frac{\frac{T}{p} \sqrt{1-p}}{x} \ge 0.99
\end{equation}

In other words, values of $T$ and $p$ which meet the requirements expressed in Equations \ref{eq:xplanned} and \ref{eq:xplannedmargin}, will produce an RI with exponential distribution of inter-assignment intervals that is sufficiently close to a RI$x$ (of same size) as planned beforehand. A small R script to determine adequate combinations of $T$ and $p$ is available as supplemental material (Appendix~\ref{ap:T_p}).

After choosing appropriate values for $T$ and $p$, the simulation starts running. A given interval will elapse until the first reinforcer is assigned. After every reinforced response, the chronometer restarts. That poses the interval schedule’s criterion for reinforcement presentation based on the time period between two consecutive reinforcers (reinforcement as a function of both responding and passage of time). Using such an implementation, based on \citet{Millenson1963}, we will discuss the shape of the RFF RI produced using Beak. 

\subsubsection{Random Differential Reinforcement of Low Rates (RDRL)}

In the well-known DRL schedule (differential reinforcement of low rates of behavior), a minimum inter-response time (IRT) must precede rewarded responses \citep{Ferster1957}. Using Beak, we were able to implement the variable differential reinforcement of low rates - the RDRL schedule \citep{Ambler1973, Logan1967}. In a RDRL schedule, the required IRT varies randomly. Such variation is a function of parameters like those used to implement the RI schedule \citep{Millenson1963}.

Just like the previously defined RI, a reinforcement is assigned with probability equal to $p$ every $T$ seconds. The difference relies on the fact that, in the RI schedule, the parameter $T$ is not affected by the organism's behavior, whereas the same parameter, in the RDRL, is directly affected by the organism’s IRT. This happens because the chronometer that registers the passage for each cycle resets after every response emitted, which causes a cycle of time $T$ to only be fully completed if no responses are emitted in the meantime. Such a condition makes $p$ conditional to the organism’s IRT, so in order to obtain a mean value for the probability of reinforcement in the session one must consider the minimum IRT the schedule requires (the size of the RDRL).

In other words, while the relation between $T$ and $p$ defines the average IAT of an RI, the same relation defines the average IRT which the organism is required to comply with in order to produce reinforcers in a RDRL. Therefore, substituting $T$ for $T'$, in order to emphasize such a difference, the mean RDRL size is given by: 

\begin{equation}\label{eq:muRDRL}
\mu_{RDRL} = \frac{T'}{p}
\end{equation}

The parameter $T'$ is the minimum IRT required by the schedule for reinforcement assignment and $p$ is the probability that a reinforcer is actually assigned by the end of $T'$. Here we'll use Beak to draw the RFF RDRL and discuss a convenient curve fit. Even though the RDRL was implemented in animal laboratory \citep{Logan1967, Aasvee2015}, to the best of our knowledge, no further studies have been published about the RFF RDRL. Therefore, we will discuss this matter in the section in which we cover the advances we were able to make using Beak. 

\subsubsection{Simulating responses}

Here we will present the assumptions of Beak regarding the implementation of responses to study schedules of reinforcement using computational simulation. Beak produces instantaneous responses programmed as a Bernoulli process, where a success corresponds to the emission of a response. The simulation explores a range of response rates ($B$), being $B$ constant along each session. The probability of response emission at each instant of time ($p_b$) for each session is given by:

\begin{equation}\label{eq:probbeak}
p_b = \frac{B}{60 \frac{1}{t}}
\end{equation}

The simulation evolves in discrete steps. Each second is fractioned according to $t$ (the minimum possible IRT). The mean rate of responses, $B$, is provided in minutes (the correspondence from minutes to seconds is represented by the constant 1/60 in Equation \ref{eq:probbeak}). For instance, a response rate of 100 per minute and a second partitioned in intervals of 5/1000 of a second, would result in $p_b \approx 0.0083$ (the probability of response in each iteration step). To investigate higher values of $B$, $t$ needs to be smaller, making the simulation finer with greater computational cost. Additionally, as this rate of trials increases, the Bernoulli process approaches a time continuity, as in a Poisson process.

The researcher also determines session duration and the number of session repetitions. Beak stores the reinforcement rate (reinforcers per minute) of each repetition and computes the 95\% highest density interval \citep{Hyndman1996}. For this work we computed 500 repetitions of one-hour sessions, therefore, each point of our simulations corresponding to a given $B$ was the result of 500 sessions, each one depending on $720,000$ iterations ($3600 \cdot \frac{1}{0.005}$), totaling $3.6 \cdot 10^8$ trials. Since $B$ ranged from 0 to 200 (integer values), the definition of each RFF depicted below was obtained by $7.236 \cdot 10^{10}$ events. With such a number of trials, the obtained average rate of responses draws itself nearer to the nominal rate of responses determined by the experimenter.

\subsection{Break-and-run patterns}

For another simulation set we have implemented two new parameters in order to implement a break-and-run pattern of responses \citep{Nevin1980}. We have started these simulations running a probability of starting a new burst of responses (i.e., probability of a run, $P_r$), and during a run, a probability of stopping the emission of responses (i.e., the probability of a break, $P_b$). Responses emitted during this burst vary around a mean Local Operant Rate (LOR) according to a Bernoulli process. Such simulations allow us to compare two different accounts about the structure of behavior, one that assumes that behavior is random and other that assumes clusters of behavior that start and stop randomly. 

\section{Results}

Our results include a comparison of different RFF for the RI, and a possible RFF for the RDRL. For both schedules, in addition to graphic representations, we consider how well each RFF fits our simulated data using a goodness of fit measure ($R^2$). Also, for RI schedule, we computed the Bayesian information criterion (BIC) and Akaike information criterion (AIC) values as parsimony and generality measures.

\subsection{RI curve fit}

\begin{figure}[ht]
\begin{center}
\includegraphics[width=5in]{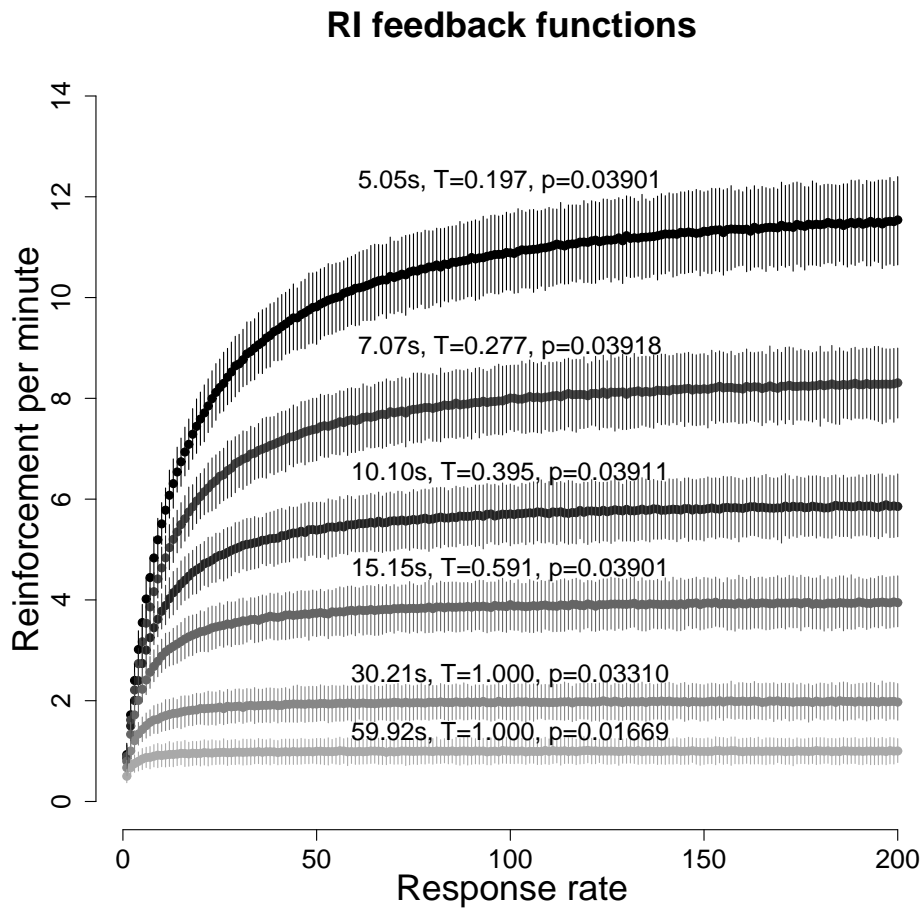}
\end{center}
\caption{Simulated data relating mean reinforcement per minute with responses per minute in three RI
schedules applying the best approximation of RI~5s, RI~7s, RI~10s, RI~15s, RI~30s and RI~60s with IAT geometrical distribution as function of $T$ and $p$ (see text). Average is represented by points and 95\% high density intervals by vertical bars from 500 repetitions of simulated sessions of 1 hour.}\label{fig:feedbackRI}
\end{figure}

The results of our simulation are in agreement with \citet{Millenson1963} RI findings. We also found that the rate of reinforcement in RI schedules is monotonically increased and negatively accelerated, as shown in Figure~\ref{fig:feedbackRI}. 

Ahead we will compare different RFF for the RI. We have tested RFF by \citet{Baum1981}, \citet{Prelec1982}, and \citet{Rachlin1978} equations. We also tested a new RFF for the RI, presented in a different theoretical context by \citet{Killeen1975}. These RFF are summarized in Table~\ref{tab:RI}. In all RFF, $B$ stands for the response rate, $R$ stands for reinforcement rate and $V$ stands for the size schedule in minutes, while $c$ and $m$ are free parameters that are estimated \textit{a posteriori}.

\begin{table}[H]
\caption{Equations explored herein investigating best RFF RI fit ($V$ provided in seconds and scaled by 60 for conversion in minutes).}\label{tab:RI}
\begin{center}  
\begin{tabular}{l c} 
\toprule
 \textbf{Reference} & \textbf{RFF RI}
\\
\midrule
\citet{Baum1981} & ${R = {\frac{1}{(V/60)+\frac{1}{B}}}}$ \\  
\vspace{0.1in}
\citet{Killeen1975} & ${R = { \frac{1}{(V/60)} \left( {1 - exp \left( -\frac{B}{c}\right)} \right) }}$\\  
\vspace{0.1in}
\citet{Prelec1982} & ${R = {B \left( {1 - exp \left( -\frac{1}{(V/60)B}\right)} \right) }}$\\  
\vspace{0.1in}
\citet{Rachlin1978} & ${R = {\frac{1}{(V/60)} \left( \frac{B}{B_{max}} \right)^m}}$\\  
\bottomrule
\end{tabular}  
\end{center}
\end{table}

To fit the curves presented to the data we obtained through simulation, we allowed all parameters of the equations to vary (except $B$ and $R$ which are the variables we want to describe) in order to best fit the curve to the data by a nonlinear least squares method. From that ensues that we have estimated parameters, which are not exactly those obtained empirically but are a good approximation. For instance, we have an estimated $V$ that approaches the schedule's size that was defined by the experimenter but renders a more accurate description of the data than that one defined \textit{a priori}. We do that because we want to know how faithfully one can assume that this parameter actually approaches the schedule’s size. Therefore, we investigated how $V$, $m$ and $c$ vary across RI. These results are summarized in Table~\ref{tab:RIbeak}.

\begin{table}[H]
\caption{Parameter estimates for each RFF RI using Beak}\label{tab:RIbeak}
\begin{center}  
\begin{tabular}{cc|cccccccc}
\toprule
& \textbf{RFF} & \textbf{Parameter} & \textbf{RI5} & \textbf{RI7} & \textbf{RI10} & \textbf{RI15} & \textbf{RI30} & \textbf{RI60} & 
\\
\midrule
& \citet{Baum1981} & $V$ & 4.91 & 6.92 & 9.92 & 14.88 & 29.86 & 59.56 & \\  
\midrule
& \citet{Killeen1975} & $V$ & 5.41 & 7.49 & 10.55 & 15.59 & 30.72 & 60.58 & \\ 
&  & $c$ & 18.474 & 14.059 & 10.441 & 7.383 & 3.961 & 2.107 & \\  
\midrule
& \citet{Prelec1982} & $V$ & 5.25 & 7.29 & 10.33 & 15.32 & 30.38 & 60.17 & \\  
\midrule
& \citet{Rachlin1978} & $V$ & 4.90 & 6.83 & 9.71 & 14.52 & 29.20 & 58.60 & \\  
&                  & $m$ & 0.210 & 0.174 & 0.142 & 0.111 & 0.071 & 0.043 & \\  
\bottomrule
\end{tabular}  
\end{center}
\end{table}  

Table \ref{tab:RIbeak} shows that our estimations of $V$ are all fairly close to the chosen schedule sizes. The parameter $m$ \citep{Rachlin1978, Rachlin1989} seems to be an inverse function of the RI size. The parameter $c$ \citep{Killeen1975} shows a similar behavior across RI sizes, but it does not seem to have an upper limit. 

As previously mentioned, an appropriate feedback function should fit the data \citep{Baum1992}. In order to compare fit qualities, one possible criteria is the goodness of fit measure, $R^2$, for what we suggest the threshold 0.9 and 0.95 for good and excellent fit, respectively. Notwithstanding, using  $R^2$ as the only criterion could be misleading, since it usually favors more complex RFF. Hence, we will use BIC and AIC to compare models with different numbers of parameters \citep{Schwarz1978}. Table \ref{tab:RIfit} summarizes the $R^2$ and BIC estimated for each RFF.

\begin{table}
\caption{Fit precision and parsimony for each feedback function using simulated data from three RI schedules}\label{tab:RIfit}
\begin{center}  
\resizebox{\textwidth}{!}
{
\begin{tabular}{cc|crrr|crrrc} 
\toprule
& \textbf{RFF}        & \textbf{RI}   & $R^2$    & \textbf{BIC}   & \textbf{AIC}      & \textbf{RI}   & $R^2$    & \textbf{BIC}   & \textbf{AIC} & \\
\midrule
& \citet{Baum1981}    & 5s  & 0.99993 & -1070 & -1077  & 7s  & 0.99985 & -1101 & -1107 & \\
& \citet{Killeen1975} &     & 0.96347 &   189 &   179  &     & 0.95274 &    62 &    52 & \\
& \citet{Prelec1982}  &     & 0.93349 &   304 &   297  &     & 0.92918 &   137 &   131 & \\
& \citet{Rachlin1978} &     & 0.88794 &   414 &   404  &     & 0.85478 &   286 &   276 & \\
\midrule
& \citet{Baum1981}    & 10s & 0.99975 & -1185 & -1192 & 15s & 0.99933 & -1220 & -1226 & \\
& \citet{Killeen1975} &     & 0.93744 &   -77 &   -87 &     & 0.92245 &  -263 &  -273 & \\
& \citet{Prelec1982}  &     & 0.92353 &   -43 &   -49 &     & 0.91713 &  -255 &  -261 & \\
& \citet{Rachlin1978} &     & 0.82013 &   134 &   124 &     & 0.76983 &   -45 &   -55 &  \\
\midrule
& \citet{Baum1981}    & 30s & 0.99696 & -1330 & -1337 & 60s & 0.98758 & -1485 & -1492 & \\
& \citet{Killeen1975} &     & 0.88656 &  -601 &  -611 &     & 0.83498 &  -963 &  -973 & \\
& \citet{Prelec1982}  &     & 0.89932 &  -630 &  -637 &     & 0.86361 & -1006 & -1013 & \\
& \citet{Rachlin1978} &     & 0.68004 &  -394 &  -404 &     & 0.58907 &  -780 &  -790 &  \\
\bottomrule
\end{tabular}  
}
\end{center}
\end{table}

Our results favor \citeauthor{Baum1981}'s \citeyearpar{Baum1981} RFF regarding both excellent fitting (highest $R^2$) and parsimony (lowest BIC). Overall, the $R^2$ seems to decay as the RI sizes increase.

\begin{figure}[H]
\begin{center}
\includegraphics[width=4.5in]{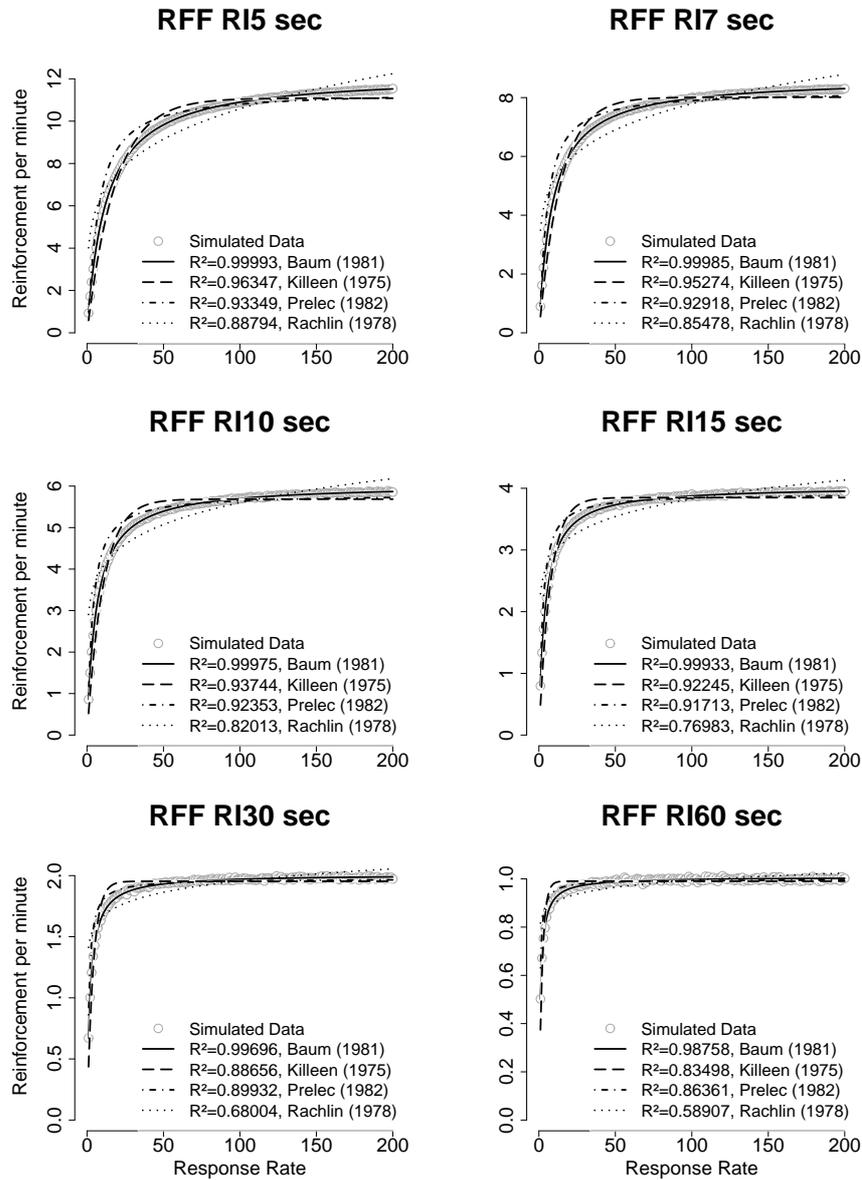}
\end{center}
\caption{Curve fit and $R^2$ for each feedback function using simulated data of RI 5s, 15s, and 60s.}\label{fig:feedbackRIfit}
\end{figure}

\clearpage
Figure \ref{fig:feedbackRIfit} brings graphical representation for the RI 5s, 15s, and 60 seconds and helps to understand how this functions fit our simulated data.

\subsection{RDRL Feedback Function}

\begin{figure}[ht]
\begin{center}
\includegraphics[width=4in]{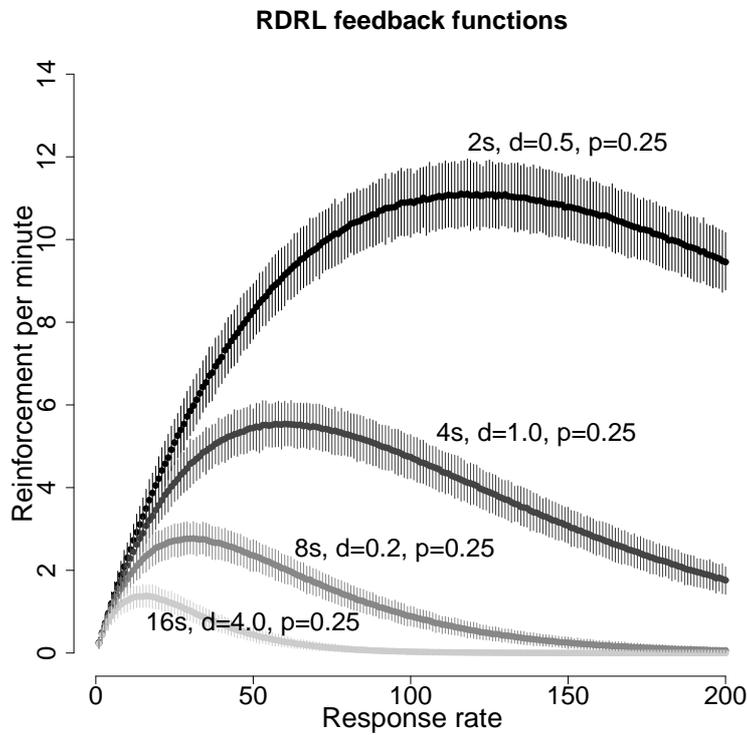}
\end{center}
\caption{Simulated data relating reinforcement per minute with responses per minute in four RDRL schedules.}\label{fig:feedbackRDRL}
\end{figure}

As described, a RDRL could reinforce any IRT with a certain probability. Therefore, we expect an optimal rate greater than the size of RDRL and, as a result, a maximum of reinforcers per minute falls short of the theoretical asymptote deduced from the size of schedule. All these features are shown in Figure \ref{fig:feedbackRDRL}, which depicts the points resulting from our simulation of four different RDRL, namely 2s, 4s, 8s and 16s. 

Our simulations are well described by the equation:

\begin{equation}\label{eq:RDRL}
R = { a \left( exp \left( -\frac{B}{b} \right) - exp \left( -\frac{B}{c} \right) \right)  }
\end{equation}

As for the RI schedule, $R$ and $B$ stand for rates of reinforcement and responses, respectively. Also, $V$ still stands for the schedule size in seconds. The parameter $a = \frac{1}{V/60} = \frac{60}{V}$ is a theoretical asymptote of reinforcers per minute, a constant for which not estimative is required. 

\begin{figure}[H]
\begin{center}
\includegraphics[width=5.8in]{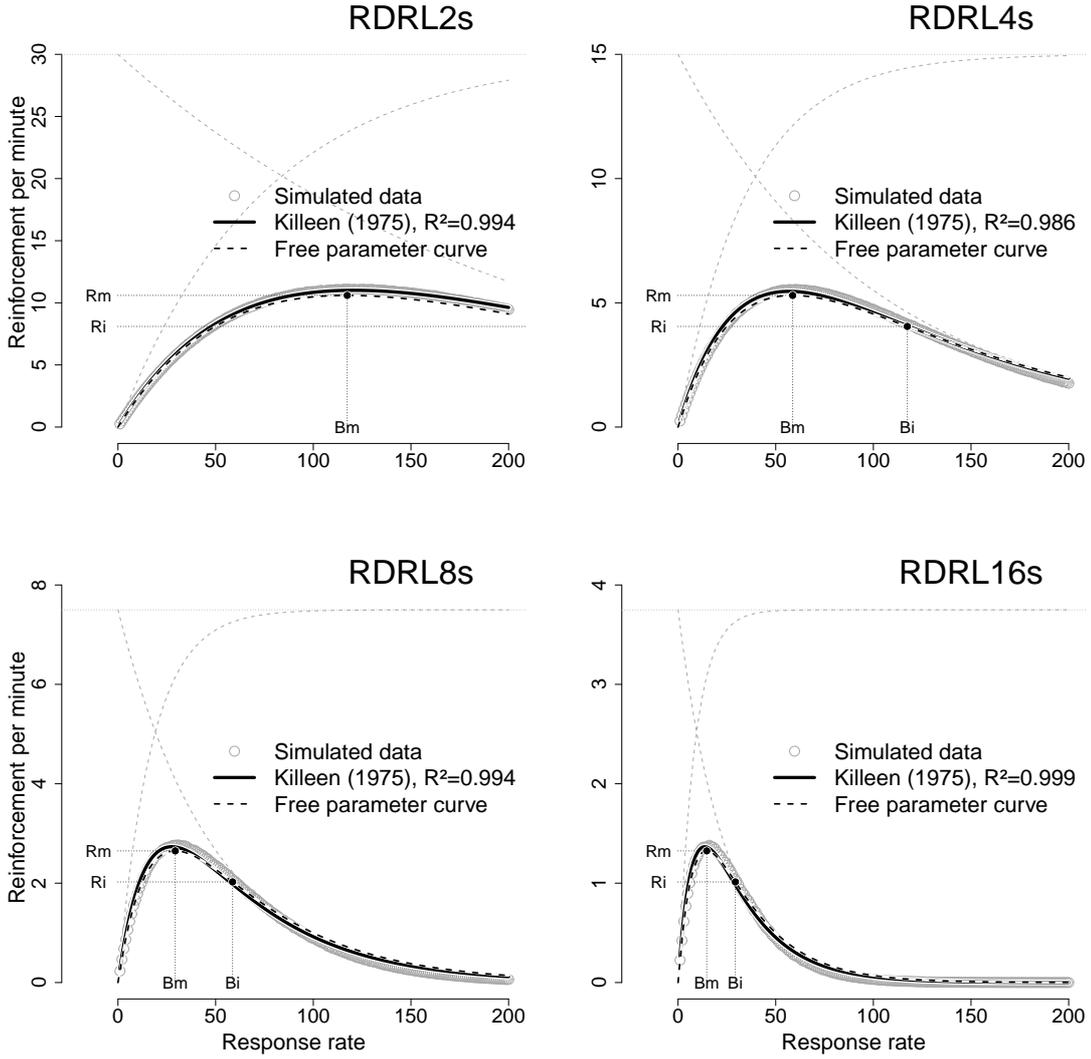}
\end{center}
\caption{Curve fit for simulated data of RDRL of sizes $V$. Top horizontal dashed line is $\frac{60}{V}$, decreasing dashed lines are adjusted by $\frac{60}{V} exp\left(-\frac{B}{b}\right)$, rising dashed lines are adjusted by $\frac{60}{V} \left(1-exp\left(-\frac{B}{b}\right)\right)$, and empty circles are simulated data. Solid tick line is RDRL fit by ${\frac{60}{V} \left(exp\left(-\frac{B}{b}\right)-exp\left(-\frac{B}{b}\right) \right)}$, and dashed lines is provided by ${\frac{60}{V} \left( exp\left(-\frac{V}{e^{6}} B\right) - exp\left(-\frac{V}{e^{5}} B\right) \right)}$, where $B$ is the response rate, $(Bm,Rm)$ is the maximum point, and $(Bi,Ri)$ is the inflection point.}\label{fig:RDRLfit}
\end{figure}

Using an iterative least squares algorithm, we have estimated the parameters $b$ and $c$ (Table \ref{tab:RDRL}) for all RDRL in Figure \ref{fig:feedbackRDRL}. The $R^2$ values summarized in this table show that Equation \ref{eq:RDRL} is a proper RFF for the RDRL schedule (we dismiss a Bayesian Information Criterion analysis simply because we do not know any viable alternative to model the RDRL). For further discussion of these parameters in which $b$ controls the decreasing and $c$ the increasing of obtained reinforcements, figure \ref{fig:RDRLfit} exemplifies this relation, comparing 2, 4, 8, and 16 seconds RDRL.

\begin{table}[H]
\caption{Parameter estimates ($b$ and $c$) and $R^2$ for simulated data of four RDRL schedules. $B$ is response rate and $V$ is schedule size (seconds).}\label{tab:RDRL}
\begin{center}  
\begin{tabular}{cc|rrrr} 
\toprule
 &  & \multicolumn{4}{l}{$a \left( exp \left( -\frac{B}{b} \right) - exp \left( -\frac{B}{c} \right) \right)$}  \\
& $V$        & $a=\frac{60}{V}$   & $b$   & $c$  & $R^2$ \\
\midrule
& 2s & 30.00 & 212.55 & 74.94 & 0.994  \\
& 4s & 15.00 & 99.68 & 35.46 & 0.986  \\
& 8s & 7.50 & 48.25 & 17.14 & 0.994  \\
& 16s & 3.75 & 23.94 & 8.51 & 0.999  \\
\bottomrule
\end{tabular}  
\end{center}
\end{table}

However, it is also noticeable that $b$ and $c$ vary in a regular proportion. By assuming $c = \frac{b}{e}$, we were able to reduce equation~\ref{eq:RDRL} to a single parameter $b$. In addition to that, it is also possible to show that $ln(b) = -ln(V)+6$, thus reducing equation~\ref{eq:RDRL} to equation~\ref{eq:RDRL1}, an equation with no free parameters that allows us to calculate reinforcement rate \textit{a priori}, similar to the RI RFF by \citet{Baum1981}. The comparison between data fit from models provided by equations~\ref{eq:RDRL} and \ref{eq:RDRL1} is also shown in Table~\ref{tab:RDRL}.

\begin{equation}\label{eq:RDRL1}
R = { \frac{60}{V} \left( exp\left(-\frac{V}{e^{6}} B\right) - exp\left(-\frac{V}{e^{5}} B\right) \right)  }
\end{equation}

\begin{equation}\label{eq:RDRL2}
R_{m} = { \frac{60}{V} \left( e - 1 \right) exp{ \left( - \frac{e}{e-1} \right) }  }
\end{equation}

\begin{equation}\label{eq:RDRL2b}
B_{m} = \frac{e^6}{(e-1)V}
\end{equation}

\begin{equation}\label{eq:RDRL2c}
B_{i} = \frac{2e^{6}}{(e-1)V} = 2 B_{max} 
\end{equation}

\begin{equation}\label{eq:RDRL2d}
R_{i} = \frac{60}{V} \left(exp\left(-\frac{2}{e-1}\right) - exp\left(-\frac{2e}{e-1}\right)\right)
\end{equation}

\subsection{Break-and-run influence}

\begin{figure}[H]
\begin{center}
\includegraphics[width=4in]{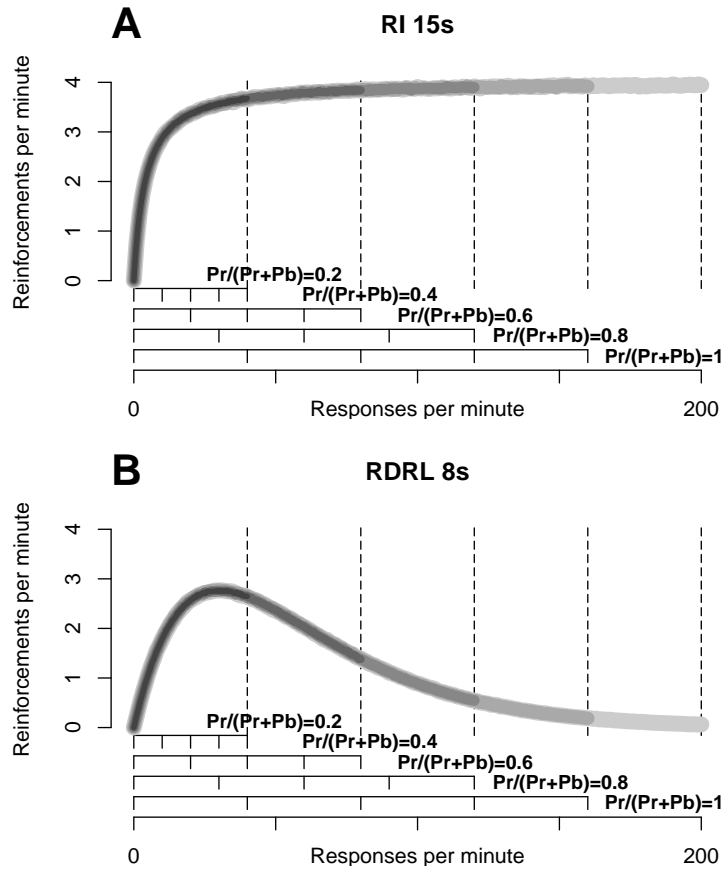}
\end{center}
\caption{Influence of burst and stop behaviors in the simulation of two exemplary schedules: (A)~RI~15s and (B)~RDRL~8s with several levels of probability to start a burst ($P_b$) and to start a pause ($P_s$). It was found that nominaly ranging from 0 to 200 pecks per minute the mapping is not affect; the curves are only incomplete for the effective number of pecks per minute is $\frac{Pb_s}{P_b+P_s}$.}\label{fig:BurstStop}
\end{figure}

It was simulated the influence of several combinations of the probabilities to start a run and a break. Contrary to some intuitive expectation, the reproduction of these break-and-run patterns did not influence the distribution of reinforcements for different rates of responses. The effective rate of pecks per minute is merely the product of the nominal pecks per minute by the proportion of time in which the animal is behaving, $\frac{P_r}{P_r+P_b}$~(Figure~\ref{fig:BurstStop}). Consequently, to obtain the map of a given effective rate in a certain schedule, it is equivalent to simulate the equivalent nominal rate without break-and-run (i.e., $P_r=1$ and $P_b=0$). 

\clearpage
\section{Discussion}

The main objective of the present paper was to implement and discuss simple schedules of reinforcement from a quantitative perspective. In Beak, we have implemented random interval (RI) and random DRL (RDRL) schedules. Also, we have implemented a random distribution of responses that allow us to investigate wide ranges of response rate and their effects on reinforcers per minute (i.e., RFF). It is important to emphasize that these simulations do not replace the study of animal behavior. Simulations are concerned with mapping of an entire schedule, going through a large range of possible response rates and exhaustively repeating these conditions. In this sense, Beak can provide orientation for a researcher in creating an experimental scenario to which a biological being can be purposefully subjected. Since this biological being will behave with certain response rate, its confrontation with the simulation predictions may clarify biases and constraints of actual behavior. In other words, simulations map the normative rules of schedules while experiments map effective behaviors of organisms.

Comparisons between different RFF for RI provide experimenters with better ways to describe the relationship between behavior and environmental constraints. However, deciding between curve fits is no simple matter, given that there are no definitive criteria. We will address the issue systematically, highlighting the pros and cons of each one of the four curve fits - \citet{Baum1981, Prelec1982, Rachlin1978, Killeen1975}.

\subsection{RI RFF} 

As Figure~\ref{fig:feedbackRI} shows, RI simulations executed using Beak were able to reproduce the general shape of a known RFF. That enables the investigation of how precisely and parsimoniously the RFF curves proposed can describe simulated data, which gives us grounds to point out advantages and disadvantages of different ways to implement each RFF.

\subsubsection{Meaning}
To describe the relationship between $B$ and $R$, \citeauthor{Baum1981}'s \citeyearpar{Baum1981} and \citeauthor{Prelec1982}'s \citeyearpar{Prelec1982} RFF rely only on $V$, a single parameter which has a built-in meaning and is supposed to correspond to the schedule’s size determined from experimental planning. That is convenient, especially because, at least in theory, they do not require estimation methods. On the other hand, \citeauthor{Rachlin1978}'s \citeyearpar{Rachlin1978} depends on $m$, and \citeauthor{Killeen1975}'s \citeyearpar{Killeen1975} depends on $c$. As far as we know, these parameters have no empirical meaning.

\citet{Rachlin1978, Rachlin1989} showed that $m$ always falls between zero and one for any interval schedule and suggested that $m=0.1$  \citep{Rachlin1989} or $m=0.2$ \citep{Rachlin1978} across schedules of different sizes. In fact, if $m$ approaches zero, the interval schedule approaches a RT; if $m$ approaches one, it approaches a RR. However we did not found a constant value for $m$, which varies with RI size.

\begin{equation}\label{eq:RIRalt}
R  = \frac{60}{V} \left(\frac{B}{B_{max}}\right)^{exp\left(-\frac{1}{2}-\left(1-\frac{1}{e}\right) \; ln(V)\right) }
\end{equation}

\citeauthor{Killeen1975}'s \citeyearpar{Killeen1975} $c$ seems to be a positive value with no upper limit. Like \citeauthor{Rachlin1978}'s \citeyearpar{Rachlin1978} $m$, it seems to have an inverse relation with the schedule’s size. Still, there’s no obvious way to derive them \textit{a priori}. Therefore, we argue that \citeauthor{Baum1981}'s \citeyearpar{Baum1981} and \citeauthor{Prelec1982}'s \citeyearpar{Prelec1982} feedback functions have a didactic advantage, since they rely on a single and interpretable parameter $V$.

\subsubsection{Precision and Parsimony}
While the other RFF seem to struggle with larger RI, \citeauthor{Baum1981}'s \citeyearpar{Baum1981} RFF is fairly stable. Adding the fact that it has a single meaningful parameter, $V$, it seems to be the best available RFF RI. \citet{Baum1992} stated that \citeauthor{Rachlin1978}'s \citeyearpar{Rachlin1978} RFF has many fallouts. First, it does not have a horizontal asymptote, an important feature of interval schedules. Second, and more importantly, it does not fit the data properly (see Table \ref{tab:RIfit} and Figure \ref{fig:feedbackRIfit}). Even though \citeauthor{Prelec1982}'s  \citeyearpar{Prelec1982}  and  \citeauthor{Killeen1975}'s \citeyearpar{Killeen1975} do better than \citeauthor{Rachlin1978}'s \citeyearpar{Rachlin1978},  their $R^2$ also drops significantly for the larger RI. We cannot state that this represents a tendency for even larger RI, but that is further evidence that \citeauthor{Baum1981}'s \citeyearpar{Baum1981} is the best among them. 

\subsubsection{Generality}
Following the above criteria, one should readily decide in favor of \citeauthor{Baum1981}'s \citeyearpar{Baum1981} RFF. However, \citeauthor{Baum1981}'s function, like \citeauthor{Prelec1982}'s, applies only to interval schedules. Conversely, \citeauthor{Rachlin1978}'s \citeyearpar{Rachlin1978} and \citeauthor{Killeen1975}'s \citeyearpar{Killeen1975} functions also describe other simple schedules \citep[see also][]{Kileen2003}. \citeauthor{Rachlin1978}'s \citeyearpar{Rachlin1978} exponential function describes time, interval, and ratio schedules, but in this case such generality does not compensate for the fallouts we already discussed. Similarly, \citeauthor{Killeen1975}'s \citeyearpar{Killeen1975} functions have generality as the primary advantage, since it models not only interval schedules but also ratio schedules and the still undocumented RDRL feedback function, as discussed below.


\subsection{RDRL RFF}

\citet{Killeen1975} used Equation \ref{eq:RDRL} to model two competing processes controlled by parameters $b$~(concurrent) and $c$~(inhibitory). \citet{Killeen1975} interpreted the former as a measure of an increasing competition among different activities, and the latter as post-reinforcement inhibition. Despite the main concern in \citet{Killeen1975} analysis is the probability of behavior in inter-reinforcement intervals, we found an analogous conflict in our analysis. The RDRL poses a similar competition between contingency and postponement of reinforcers: on one hand, we have the negative punishment imposed by the schedule to rates above the schedule's criterion (controlled by $b$), on the other, we have the direct relation between rates of response and reinforcement (controlled by $c$). 

\citet{Logan1967} exposed rats to a variable DRL with only two equally likely IRT requirements. Here, we have implemented a RDRL, a continuous version of the somewhat minimalist Logan’s RDRL. Even though \citet{Logan1967} described his results in terms of proportion of IRT, the data allows us to conjecture about the RFF RDRL shape. 

Logan found that the most likely IRT observed in the experiment “approximated an optimal strategy for maximizing reward” \citep[][p.~393]{Logan1967}. This meant that the subjects’ first response after reinforcement occurred with an IRT slightly longer than the smaller RDRL interval out of the two programmed, and further responses happened with IRT around the other (longer) RDRL interval. Therefore, he found two peaks of likely IRT that matched the RDRL intervals used. 

Considering that behavior rate equals the reciprocal of IRT, \citeauthor{Logan1967}'s results allow us to sense what a RDRL RFF should look like. Reinforcers per minute should increase along with response rate until a certain maximum. However, if the response rate increases beyond this optimal point, reinforcers income would decrease asymptotically. Since \citet{Logan1967} built his RDRL out of two intervals, optimal rates of response were predictable. In fact, rats that served as subjects learned how to maximize reinforcers responding after the shorter interval and then waiting for the longer one.
However, using a geometric distribution for the values of the schedule we should expect the peaks observed in Logan's experiment to merge, forming the curves seen in Figures 4 and 5. Figure 5 shows that the greater the size of the schedule programmed the sharpest the curve on the peak of the RFF.
Another information we take from the simulations and the estimations of the RDRL RFF is about the maximum reinforcement rate possible to be obtained. Equation 9 shows the relation between maximization and the size of the schedule. An interesting result we observed across many simulations is that the maximum reinforcement rate is a constant $\frac{60}{e \; V}=0.37 a$. This normative rule not only confers \textit{a priori} the maximum reinforcement an animal can obtain given the schedule size but also the prediction of the response rate at which this maximum will be achieved.
Equation 10, in contrast, gives you the maximum rate of behavior that can be emitted in a RDRL to maximize reinforcements as a function of the schedule size. This is also interesting for planning experiments when you know the typical LOR of the subjects.

The decreasing dashed lines for each RDRL in Figure \ref{fig:RDRLfit}  are controlled by the parameter $b$, while the rising dashed lines are controlled by $c$. Greater values of $b$ mean a slower decay in reinforcer income at higher rates of response. Greater values of $c$ mean a slower increase in reinforcement at low rates of response. 

As shown in Table \ref{tab:RDRL}, we’ve found greater values of $b$ and $c$ for smaller (richer) RDRL. That matches our interpretation of the model, because smaller RDRL are more demanding and permissive. They are demanding because they require greater response rates in order to reach maximum reinforcement income, and they are permissive because they allow greater response rates to go unpunished (see Figure \ref{fig:RDRLfit}).

Although Beak allows testing for more extreme schedules, for both RI and RDRL simulations are not feasible for the response rate range presented here. For instance, a very rich schedule such as RI~0.1s up to 200 pecks per minute would barely start its ascension; conversely, a very poor schedule such as RI~600s would be already at the maximum plateau with 1 peck per minute. These schedules would require, respectively the study of responses rates up to several hundreds pecks per minute, or the study of fractional intervals in response rates between 0 and 1. This kind of resulting curve would be qualitatively equal to the ones presented here, but not applicable to emulate animal behavior, at least not for the usual animals available in our laboratories, and not contributing for the testing of curve fitting.

These findings are important because they successfully add complexity in our basic description of simple schedules of reinforcement. This complexity may be viewed as consistent with other schedule parameters, showing the capacity of our computational model to compare different quantitative models of simple schedules of reinforcement as a starting point to analysis of other sources of control, including conflict between excitatory and inhibitory control \citep{Staddon1977} and temporal control \citep[e.g.,][]{Machado1997}.

Typical behavior of a real animal might present break-and-run patterns, with bursts of responding at approximately constant tempo, and pauses of variable intervals. It is arguable that the structure of behavior in the real organism is more complex than that implemented in this model~\citep{Nevin1980}, thus more realism might be incorporated to make this model more useful for more specific occasions. In order to assess the impact of this the break-and-run model, it was necessary to implement this break-and-run feature in our simulation. Somewhat against intuition, our results show that the effect of this modification merely reproduced the behavior of a lower response rate, thus it may be removed from the model for the sake of parsimony. More realism may introduce unnecessary complications with no gain in explanatory power.

Regarding schedules in which reinforcement may depend both on the passage of time and the occurrence of responses, the RDRL is a way to further constrain reinforcement in comparison to the RI schedule. The RFF RDRL is like RFF RI, in a sense that in both cases the rate of reinforcement depends on the response rate. Therefore, we have found increasing functions at low rates of response. However, these functions are also negatively accelerated functions. This represents the restriction imposed by time, which is present in both schedules.

The RFF of both schedules differ in the extent to which the RDRL schedules further constraints reinforcement. In the interval schedule the rate of responses has a positive monotonic relation with the ever-increasing rates of reinforcement. That is not the case in the RDRL. In the RDRL schedule, high rates of response are negatively punished by the postponement of reinforcement. In fact, this feedback system is well described by two competing processes \citep{Killeen1975}. 

Briefly, our results demonstrate the power of our computational simulation to discuss basic schedules of reinforcement and refine ways to implement them. The enormous computational power available today should be used to offer, for
instance, a variety of intervals instead of a simple shuffle of a small set of intervals
mimicking older devices. Also, based on our results, we revised RI RFF and proposed a RDRL RFF. Using computer simulation prevents unnecessary use of time and subjection of living beings in a long experimentation without a clear notion of the normative rule that may be governing the strategic options involved. The new implementation methods presented paves way for a richer study of schedules of reinforcement and their normative maximization rules, serving also as a guide towards promising questions which future experiments may want to explore.

\clearpage

\bibliography{references} 
\bibliographystyle{apasoft}

\clearpage
\appendix

\section{Algorithm to find T and p (R script)}\label{ap:T_p}

Implemented by \textbf{RI\_Txp\_paper.R}:
{\singlespace \verbinput{RI_Txp_paper.R}}

\clearpage
\section{R script for Figure~\ref{fig:feedbackRI}}\label{ap:RIraw}

Implemented by \textbf{FigJeabSimpleRI.R}:

{\singlespace \verbinput{FigJeabSimpleRI.R}}

\clearpage
\section{R script for Figure~\ref{fig:feedbackRIfit} and Tables~\ref{tab:RIbeak} and \ref{tab:RIfit}}\label{ap:RIfit}

Implemented by \textbf{FigJeabCurveFitRI.R}:

{\singlespace \verbinput{FigJeabCurveFitRI.R}}

\clearpage
\section{R script for Figure~\ref{fig:feedbackRDRL}}\label{ap:RDRLraw}

Implemented by \textbf{FigJeabSimpleRDRL.R}:

{\singlespace \verbinput{FigJeabSimpleRDRL.R}}

\clearpage
\section{R script for Figure~\ref{fig:RDRLfit} and Table~\ref{tab:RDRL}}\label{ap:RDRLfit}

Implemented by \textbf{FigJeabFitRDRL.R}:

{\singlespace \verbinput{FigJeabFitRDRL.R}}

\clearpage
\section{R script for Figure~\ref{fig:BurstStop}}\label{ap:BreakRun}

Implemented by \textbf{FigJeabBreakRun.R}:

{\singlespace \verbinput{FigJeabBreakRun.R}}

\end{document}